\def\year{2018}\relax
\documentclass[letterpaper]{article} 
\usepackage{aaai18}  
\usepackage{times}  
\usepackage{helvet}  
\usepackage{courier}  
\usepackage{url}  
\usepackage{graphicx}  
\begin{document}
%
\title{Region-based Quality Estimation Network for Large-scale Person Re-identification}
\author{Guanglu Song, \textsuperscript{1,3}\thanks{This work is done when Guanglu Song is an intern at SenseTime Group Limited} Biao Leng,\textsuperscript{1} Yu Liu,\textsuperscript{2} Congrui Hetang,\textsuperscript{1} Shaofan Cai \textsuperscript{1}\\
\textsuperscript{1} School of Computer Science and Engineering, Beihang University, Beijing 100191, China\\
\textsuperscript{2} The Chinese University of Hong Kong, Hong Kong\\
\textsuperscript{3} SenseTime Group Limited\\
\textsuperscript{1}\{guanglusong, lengbiao, caishaofan, hetangcongrui\}@buaa.edu.cn, \textsuperscript{2}yuliu@ee.cuhk.edu.hk
}

\maketitle
\begin{abstract}
One of the major restrictions on the performance of video-based person re-id is partial noise caused by occlusion, blur and illumination. Since different spatial regions of a single frame have various quality, and the quality of the same region also varies across frames in a tracklet, a good way to address the problem is to effectively aggregate complementary information from all frames in a sequence, using better regions from other frames to compensate the influence of an image region with poor quality. To achieve this, we propose a novel Region-based Quality Estimation Network (RQEN), in which an ingenious training mechanism enables the effective learning to extract the complementary region-based information between different frames. Compared with other feature extraction methods, we achieved comparable results of 91.8\%, 77.1\% and 77.83\% on the PRID 2011, iLIDS-VID and MARS, respectively.

In addition, to alleviate the lack of clean large-scale person re-id datasets for the community, this paper also contributes a new high-quality dataset, named ``Labeled Pedestrian in the Wild (LPW)'' which contains 7,694 tracklets with over 590,000 images. Despite its relatively large scale, the annotations also possess high cleanliness. Moreover, it's more challenging in the following aspects: the age of characters varies from childhood to elderhood; the postures of people are diverse, including running and cycling in addition to the normal walking state.
\end{abstract}

\section{Introduction}
\begin{figure}
\centering
\includegraphics[width=1\linewidth]{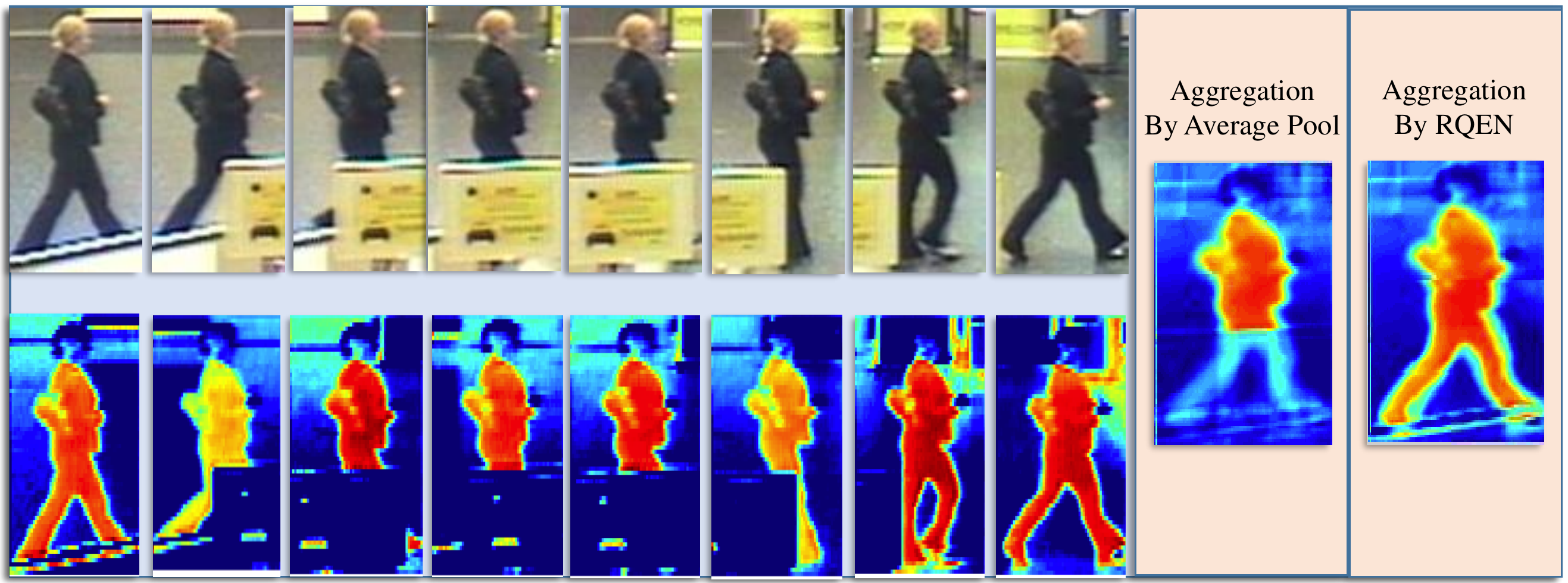}
   \caption{Illustration of different aggregation methods. The first line is an image sequence of pedestrian and the second line is the feature map extracted from the first convolution layer.}
\label{fig:feat}
\end{figure}
The purpose of person re-identification is to identify a person by comparing the similar images between the probe and a gallery set. Previous works can be divided into two categories, one is to learn a better feature extractor and the other is to design metric learning methods. Based on large-scale training data, these two types of work can achieve good performance on standard benchmarks through convolutional neural networks and carefully designed optimization strategy but may be strongly affected by occlusion or significant body movement.

In recent methods, instead of one single image, the image set of each person is used for feature extraction~\cite{mclaughlin2016recurrent}and distance metric learning~\cite{you2016top}. Compared to a single image, frames from a sequence provide richer information complementary to each other~\cite{zheng2016mars}. One intuitive way to aggregate information from a sequence is to simply take the average~\cite{karanam2015sparse}, but this can draw undesired noise into the feature. As shown in Figure~\ref{fig:feat}, some images in this sequence exemplify an occasion of partial occlusion and using the simple average method will cause recognition failure because the effective information is weakened by noise information. Recent work~\cite{yu2017quality} has proposed a novel model which generates the adaptive score for each image in a sequence. However, this score doesn't specify which part causes major limitation, and possibly leads to the dumping of a frame due to its small noisy part, causing the loss of valuable information from all other regions of that frame. Our proposed RQEN method can concentrate more attention on the effective images' regions in a sequence and aggregate the complementary region-based information between different frames.

Considering the factors above, to make our system robust to partially occluded or noisy images and selectively use complementary region-based information in the sequence, we propose a more adaptive region-based quality Estimation network (RQEN) consisting of regional feature generation part and region-based quality predictor part. It aims to weaken the influence of the images' regions with poor quality and take advantage of the complementary information in a sequence simultaneously. The method of jointly end-to-end training these two parts enables the network to learn to evaluate the information validity for different regions of the image. With the assist of latter set aggregation unit, images' regions with higher confidence will contribute more information to the representation of image sequence, and the images' regions with occlusion or noise will take up a lower proportion.

Another problem is that the current person re-identification datasets~\cite{wang2014person,hirzer2011person,li2014deepreid,zheng2015scalable} are flawed either in scale or cleanliness. Prevailing sizes range from 200 to 1500 identities, scanty for more extensive experiments. Relatively large datasets exist, but they possess poor cleanliness due to detection or tracking failure, which significantly undermines the performance of algorithms. Moreover, persons in most datasets are well-aligned by hand-drawn bounding boxes. But in reality, the bounding boxes detected by pedestrian detectors may undergo misalignment or part missing~\cite{zheng2015scalable}. To address these problems, it's important to introduce a large-scale dataset that is both clean and close to realistic settings.

For alleviating the scantiness of large-scale and clean datasets, We propose a new dataset named the ``Labeled Pedestrian in the Wild (LPW)''. It contains 2,731 pedestrians in three different scenes where each annotated identity is captured by from 2 to 4 cameras. The LPW features a notable scale of 7,694 tracklets with over 590,000 images as well as the cleanliness of its tracklets. It distinguishes from existing
datasets in three aspects: large scale with cleanliness, automatically detected bounding boxes and far more crowded scenes with greater age span. This dataset provides a more realistic and challenging benchmark, which facilitates the further exploration of more powerful algorithms.

To sum up, our contributions in this work are as follows:
\begin{itemize}
\item Be the first to take into account the quality of images' different regions to better aggregate complementary region-based information in a sequence, using the information of a certain image region with higher quality to make up for the same region of the other frames with poor quality.
\item Propose a pipeline of joint training multi-level features which enables the region-based quality predictor to generate proper evaluation of regional quality and advances state-of-the-art performance on iLIDS-VID and PRID 2011 for video-based person re-identification.
\item Construct a new person re-identification dataset named ``Labeled Pedestrian in the Wild'' with large scale and good cleanliness. It contains 7,694 tracklets and over
590,000 images, and can be available on \url{http://liuyu.us/dataset/lpw/index.html}.
\end{itemize}

\section{Related Works}

\begin{figure*}
\centering
\includegraphics[height=50mm]{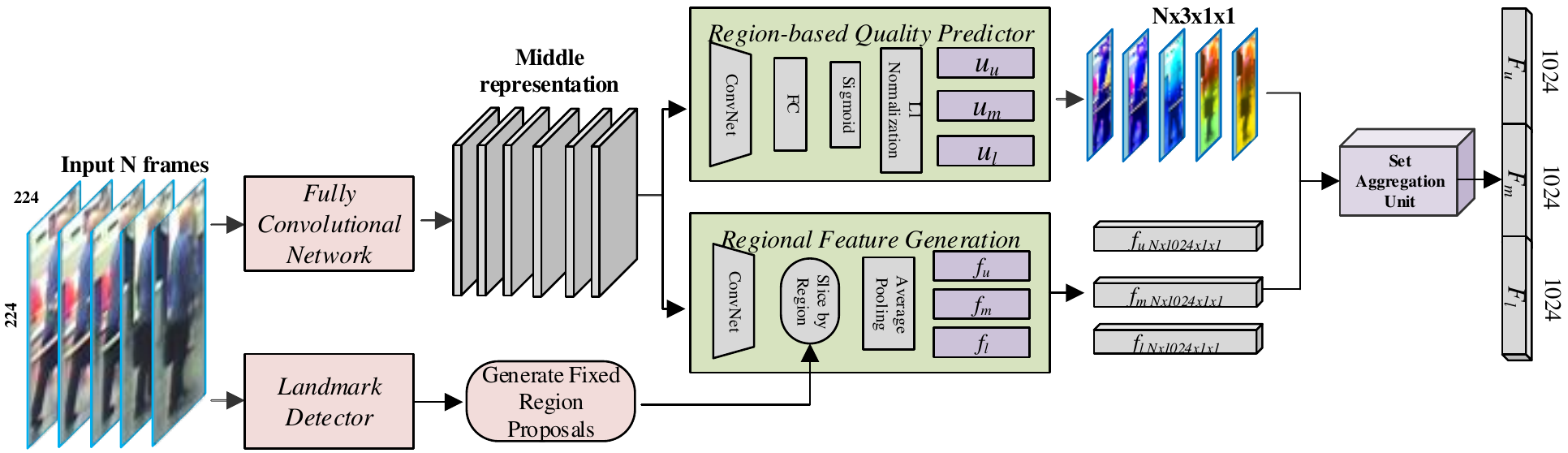}
   \caption{The inference pipline of the proposed RQEN. The input of this network is an image set belonging to the same person. Each of them generates the middle representation through the fully convolutional network. Then the representation will be fed to the regional feature generation unit with landmarks and region-based quality predictor. The scores of different regions in $[0,1]$ are indicated by the color map, i.e. from blue to red. Then the scores and features of all images will be aggregated by set aggregation unit and the final representation of the image set will be produced. $\{\mu_{u},\mu_{m},\mu_{l}\}$ and $\{f_{u},f_{m},f_{l}\}$ represent the quality scores and features of image different regions, respectively. $\{F_{u},F_{m},F_{l}\}$ means the video-level feature representation of person.}
\label{fig:pqan}
\end{figure*}
Person re-identification is crucial for smart video surveillance, but it remains unsolved due to large intra-class and inter-class variations caused by lighting change, cameras' view change, occlusion, and misalignment. Feature extraction and metric learning are the two key components that most recent works are devoted to exploring.

For feature extraction, different kinds of features~\cite{wang2007shape,liu2017rethinking} have been tailored and employed in previous work.
Cheng ~\cite{cheng2011custom} applies the drawing structure to person re-id. An adaptive body contour is used to represent the portrait, including the head, chest, thigh, and calf, and then the color characteristics of each part are extracted for matching. Zhao ~\cite{zhao2014learning} proposes the local patch matching method that learns
the mid-level filters to generate the local discriminative features for the person re-id.~\cite{wu2016enhanced} proposes Feature Fusion Net (FFN) for pedestrian image representation that
hand-crafted histogram features can be complementary to Convolutional Neural Network (CNN) features.~\cite{mclaughlin2016recurrent} uses a recurrent neural network to learn the interaction between multiple frames in a video and a Siamese network to learn the discriminative video-level features for person re-id.~\cite{yan2016person} uses the Long-Short Term Memory (LSTM) network to aggregate frame-wise person features in a
recurrent manner.~\cite{cheng2016person} uses both the global full-body and local body-parts features of the input persons to improve the person re-id performance.~\cite{xiao2016learning} learns robust feature representation with data from multiple domains for the same task. ~\cite{tn2017see} proposes an end-to-end architecture to jointly learn features and metrics.

When it comes to metric learning, some works are devoted to learning more discriminative distance metric functions. Several methods have been proposed such as KISSME~\cite{koestinger2012large}, LMNN~\cite{dikmen2010pedestrian}, Mahalnobis~\cite{roth2014mahalanobis} and applied to the person re-id problem.~\cite{you2016top} proposes the top-push distance learning model to further enhance inter-class differences, and reduce intra-class differences.~\cite{yi2014deep} uses a more general way to learn a similarity metric from image pixels directly.  FaceNet~\cite{schroff2015facenet} and Ding ~\cite{ding2015deep} apply this triplet framework to the face and person re-id task, respectively. ~\cite{li2014deepreid,ahmed2015improved} use the Siamese network architecture for feature mapping from raw images to a feature space where images from the same person are close while images from different persons are widely separated.~\cite{resystematic} proposes a systematic evaluation of person re-id task by incorporating recent advances in both feature extraction and metric learning.

All these works use either multiple images or a selected subset of a sequence to extract feature, while none of them utilize complementary information for feature representation in multiple sub-regions from different frames. Our proposed Region-based Quality Estimation Network (RQEN) introduces an end-to-end architecture that can learn the quality of different image parts and selectively aggregate complementary information of a sequence to form a discriminative video-level representation different from recent works of learning aggregation based on fixed feature~\cite{yang2016neural} or rough estimation of the image information~\cite{yu2017quality}. The corresponding features incorporating the sub-regions with high quality bear smaller intra-class variance and larger inter-class distance.

\section{Proposed Method}

\subsection{Architecture Overview}
The input of RQEN is a set $S = \{I_1,I_2,\cdots,I_n\}$ where each image belongs to the same person. At the beginning of the network, it's sent into two different parts. The fully convolutional network will be used to generate the middle representations of input images. In the regional feature generation part, the landmark detector~\cite{wei2016convolutional} marks the key points of the human body. The middle representation is divided into different regions according to the detected key points. Due to the poor resolution of people images from real-life surveillance systems, the detected landmarks for a single image usually aren't accurate enough to directly give a satisfying division. Our solution is to calculate the key point distribution of the whole dataset and figure out a generally fixed three-region-division accordingly, which approximately suits any given single image. Let $u,m,l$ represent the upper part, middle part and the lower part of images, respectively. Each part corresponds to a region of the original image and they generate feature vectors $F_{I_i} = \{f_{u}(I_i),f_{m}(I_i),f_{l}(I_i)\}$ using average pooling method with the assist of landmarks.
The other unit called region-based quality predictor can generate the quality estimation $\mu_{I_i}=\{\mu_{u}(I_i),\mu_{m}(I_i),\mu_{l}(I_i)\}$ for the corresponding regional features. On the basis of the former work~\cite{yu2017quality}, we directly feed the output feature maps of the \emph{pool2} layer into the region-based quality generation unit.
Each score of different regions $\mu_{part}(I_i)$ estimates corresponding regional feature $f_{part}(I_i)$. Eventually, the scores and their corresponding features generate the video-level feature representation $F_w(S)$ through the set aggregation unit. We now discuss these units in following sections.

\subsection{Strategy of Generating Region Proposals}
\begin{figure}[t]
\centering
\includegraphics[width=1\linewidth]{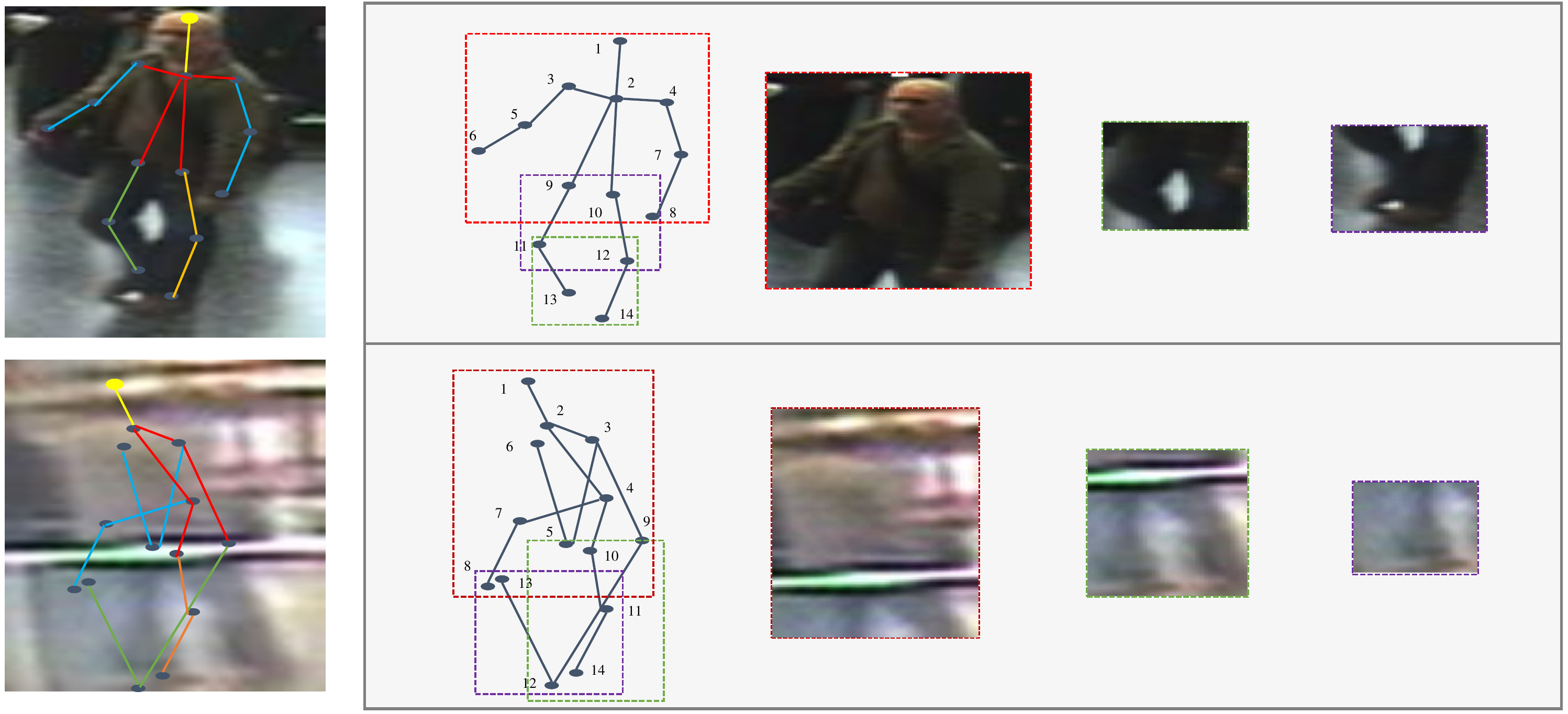}
   \caption{Samples of 14 landmarks detected on iLIDS-VID. Landmarks of the first line are detected successfully and image of the second line with low resolution and blur failed to detect.}
\label{fig:land}
\end{figure}
Let $P_{i}=\{p_{1},p_{2},\cdots,p_{m}\}$ represents the landmark set of image $I_{i}$ and $m$ is the total number of landmarks which is set 14 in our method. Figure~\ref{fig:land} shows some samples from iLIDS-VID dataset. Because of the low resolution and occlusion of some images, in many cases, landmarks are not available. A number of such detection failures exist in the datasets under surveillance. Taking into account these factors, in order to ensure more accurate generation of region proposals, we use the k-means clustering to cluster the landmarks in three sets $S_{1}^P = [1,2,3,4,5,6,7,8,9,10]$, $S_{2}^P = [9,10,11,12]$ and $S_{3}^P = [11,12,13,14]$ assigned by 14 located body joints. According to the clustering centers, the full areas of image is divided into a generally fixed three-part-division $I_{i} = \{I_{u},I_{m},I_{l}\}$ in the height direction computed by location ratio (3:2:2 in our experiment) of cluster centers. As CNN has considerable translation invariance, though regions are fixed, most body movements can be tolerated which can effectively filter landmark detection errors.

\subsection{Region-based Quality Predictor}

The quality of the image has a great influence on the feature extraction of the neural network and images of a sequence may contain a lot of noise such as blurry, occlusion and deformation. Discarding the information of the whole image because of the regional noise often leads to the loss of critical regional information which plays a key role in feature matching, so we propose the region-based quality predictor as shown in Figure~\ref{fig:pqan}. The middle representation of the input image is sent into a convolutional network consisting of convolution layer and fully connected layer. The fully connected layer generates the original scores $\mu_{ori}(I_i)$ corresponding to images' different regions and the scores are scaled to $[0,1]$ by the sigmoid function $\sigma(\cdot)$. The score of each region is generated by its corresponding feature map. In the last stage, scores belonging to the same region of the sequence are normalized, in order to facilitate video-level complementary information aggregation.

\subsection{Set Aggregation Unit}

An image set $S = \{I_1,I_2,\cdots,I_n\}$ is mapped to an representation with fixed dimension by set aggregation unit. For an image, the regional feature generation part generates the different regions' representation $F_{I_i} = \{f_{u}(I_i),f_{m}(I_i),f_{l}(I_i)\}$ and the region-based quality predictor part generates the corresponding scores $\mu_{I_i}=\{\mu_{u}(I_i),\mu_{m}(I_i),\mu_{l}(I_i)\}$. The feature representation of an image sequence can be denoted as
$F_w(S) = \{\mathcal{F}_{u}(S),\mathcal{F}_{m}(S),\mathcal{F}_{l}(S)\}$,
where $\mathcal{F}(\cdot)$ is the set aggregation function which generates feature representation with fixed dimension for a sequence with various size by incorporating all frames in a weighted manner where highly scored areas contribute more information. $\mathcal{F}(\cdot)$ is formulated as
\begin{eqnarray}
\mathcal{F}_{part}(S) = \sum_{i=1}^n\mu_{part}(I_i)f_{part}(I_i)
\end{eqnarray}
where $\mu_{part}(I_i)$ and $f_{part}(I_i)$ represent different regions' scores and features, respectively and $n$ is the number of images. All the scores are normalized, so there are $\sum_{i=1}^n\mu_{u}(I_i)=1$, $\sum_{i=1}^n\mu_{m}(I_i)=1$ and $\sum_{i=1}^n\mu_{l}(I_i)=1$.

\subsection{Learning Complementary Information Via Joint Training Frame-level and Video-level Features}

\begin{figure}
\centering
\includegraphics[width=1\linewidth]{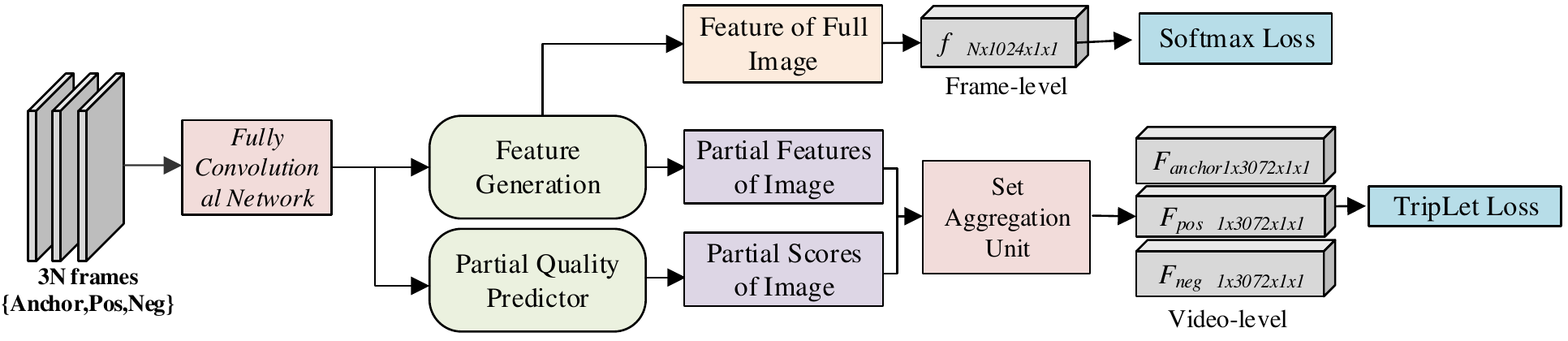}
   \caption{The training stage of RQEN. The frame-level features are supervised by Softmax loss with pedestrian identities, while video-level features are supervised by Triplet loss.}
\label{fig:train}
\end{figure}

The representation of information is not independent between different frames. The complementary information of different image regions in pedestrian image sets can effectively improve the performance of recognition. The extraction of complementary region-based information should weaken the influence of noise information and fuse more effective information of pedestrian expression. This can be accomplished via the RQEN we proposed with the assist of region-based quality scores.

Figure~\ref{fig:train} shows the training stage of RQEN.
The overall loss of RQEN can be formulated as follows:
\begin{eqnarray}
\mathcal{L} = \mathcal{L}_{softmax}+\mathcal{L}_{t}
\end{eqnarray}
Note that $\mathcal{L}_{softmax}$ is the softmax loss and $\mathcal{L}_{t}$ represents the triplet loss.
\begin{equation}
\mathcal{L}_{t}=[d(\mathcal{F}_w(S_i^o),\mathcal{F}_w(S_i^+))-d(\mathcal{F}_w(S_i^o),\mathcal{F}_w(S_i^-)) + \tau]_{+}
\end{equation}
where the $d(\cdot)$ is $L_2$-norm distances, $[\cdot]_{+}$ indicates $max(\cdot,0)$ and $\tau$ is the margin of triplet loss.
The input of the network is a triplet $S_i = < S_i^o,S_i^+,S_i^->$, and the network maps $S_i$ into a feature space with $F(S_i) = < F_w(S_i^o),F_w(S_i^+),F_w(S_i^-)>$.
We define $S_i^o$ as anchor set, $S_i^+$ as positive set, and $S_i^-$ as negative set.
In the stage of region-based quality score predictor, $T$ represents input data, $\mu_{ori}(I_i)$ stands for partial scores before normalization. The generation formulation is
\begin{eqnarray}
\mu_{ori}(I_i)=\sigma( W\cdot T + b)
\end{eqnarray}
where $\sigma(\cdot)$ is the sigmoid function scaling the scores to $[0,1]$. The stage of normalization is
$\mu_{part}(I_i)=\frac{\mu_{ori}(I_i)}{\sum_{i=1}^n\mu_{ori}(I_i)}$.

By using the supervision of pedestrian identities, the frame-level features can be closer to the pedestrian center in feature space. The video-level features can be aggregated with the assist of region-based quality scores and the supervision of triplet loss prompted it to extract effective regional information between the frame-level features to further enhance inter-class differences and reduce intra-class differences. The more robust video-level features can be extracted via joint training frame-level and video-level features.

\section{The Labeled Pedestrian in the Wild Dataset}
\begin{table*}[t]
\centering
\begin{center}
\caption{Comparing LPW with existing datasets containing MARS~\cite{zheng2016mars}, Market~\cite{zheng2015scalable}, iLIDS-VID~\cite{wang2014person}, PRID2011~\cite{hirzer2011person} and CUHK03~\cite{li2014deepreid}. The DT failure stands for whether detection or tracking failure in sequences exist. The symbol \# represents the number of corresponding aspects.}
\begin{tabular}{l|c|c|c|c|c|c}
   \hline
   Datasets & LPW & MARS & Market & iLIDS-VID & PRID2011 & CUHK03 \\
   \hline
   \#identities & 2,731 & 1,261 & 1501 & 300 & 200 & 1360 \\
   \hline
   \#BBoxes & 590,547 & 1,067,516 & 32,668 & 43,800 & 40k & 13,164 \\
   \hline
    DT failure & no & yes & yes & no & no & no \\
   \hline
   \#tracklets & 7,694 & 20,715 & - & 600 & 400 & - \\
   \hline

\end{tabular}
\label{table:dataset}
\end{center}
\end{table*}

\begin{figure}
\begin{center}
\includegraphics[width=1\linewidth]{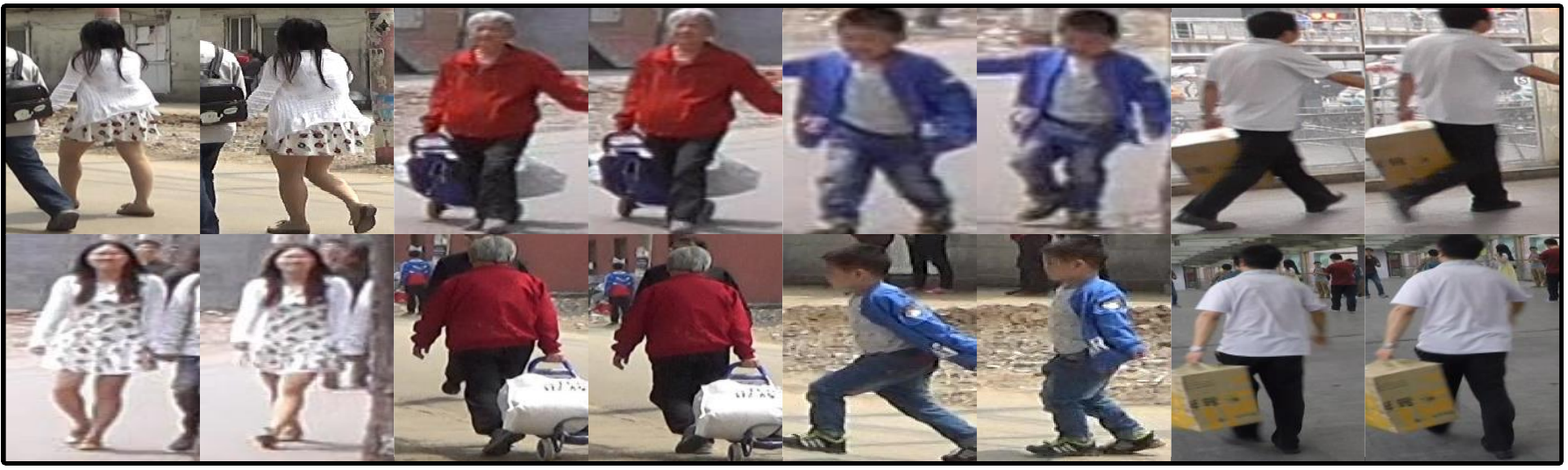}
\end{center}
   \caption{Sample images from the different cameras in the LPW. There are various postures and large span of age.}
\label{fig:sample}
\end{figure}

In this section, a new person re-id dataset, the LPW dataset, is introduced. The dataset is collected in three different crowded scenes. There are three cameras placed in the first scene and four in other two scenes. During collection, the cameras with the same parameters set were placed at the two intersections of the street. The LPW consists of 2,731 different pedestrians and we make sure that each annotated identity is captured by at least two cameras so that cross-camera search can be performed. Sample images from different perspectives of view are shown in Figure~\ref{fig:sample}.

The conversion from raw video to tracklets is threefold. First, pedestrians are captured by an elaborate detector. Then the overlaps between bounding boxes in two consecutive frames are computed to correctly assign boxed person images into their tracklets. Finally, we manually clean up the dataset, eliminating contaminated images caused by wrong detection and impure tracklets with tracking errors. As result, a total of 7,694 image sequences are generated with an average of 77 frames per sequence.
There are 1058, 1114 and 559 pedestrians captured by two, three and four cameras, respectively.

Overall, our dataset has three characteristics. First, from the comparison in the Table~\ref{table:dataset} we can see that most existing datasets are either too small in size or large in size but not clean, which negatively affects the implementation of many new ideas. Our dataset not only has a large scale in the number of pedestrians but also bears cleanliness over sequences, which meets the training requirements of many deep networks. Second, other large data sets exist in many cases where area of pedestrian account for only a small proportion of the image and in LPW the bounding boxes is fine-tuned to make sure that area of pedestrian is dominant. Third, this dataset is collected in three different crowded scenes with a large span of age, frequent occlusion, and various postures, all reflecting real-world conditions. The large-scale dataset with good cleanness can be fully exploited to obtain more discriminative information about the person of interest.


\section{Experiments}

In this section, we evaluate the performance of RQEN on two publicly available video datasets and proposed large-scale dataset for video-based person re-id: the PRID 2011 dataset~\cite{hirzer2011person}, iLIDS-VID dataset~\cite{wang2014person} ,MARS~\cite{zheng2016mars}and LPW. After that, we verify the generalization ability of our model on cross-database testing. Finally, the compatibility of generative scores with different image parts and the complementarity of information are discussed in detail.

\subsection{Datasets and Settings}

{\bf Datasets and Evaluation.} In our experiments, we adopted a multi-shot experiment setting. The evaluation of PRID 2011 and iLIDS-VID is same with CNN+RNN~\cite{mclaughlin2016recurrent}. In the LPW, the second scene and the third scene with a total of 1,975 people are used for training, and the first scene is tested with 756 people. For the LPW, the videos from $view2$ in the first scene are used as probe set and the other two views are used as the gallery set. The images of $scene2$ and $scene3$ are used for training. The reason for using $scene1$ as a test set is that the first scene contains all the challenging cases such as occlusion and significant changes of pedestrian pose in the other scenes. Features are measured using cosine distance. The results are reported in terms of Cumulative Matching Characteristics (CMC) curves. The evaluation of MARS is same as ~\cite{zheng2016mars}. To obtain statistically reliable results, we repeat the experiment ten times on PRID2011, iLIDS-VID and report the average of the results.

\begin{table}[t]
\centering

\caption{Ablation study on iLIDS-VID and PRID2011}
\scalebox{0.8}[1]{

\begin{tabular}{l|c c c c c c c}
\hline
 Conf &(a)&(b)&(c)&(d)&(e)&(f)&(g) \\
  \hline
+RU &  & $\surd$ &  & & & &$\surd$\\
+RM &  & & $\surd$ & & & &$\surd$\\
+RL&  & &  & $\surd$& & &$\surd$\\
+QFix&  & &  & & $\surd$& & \\
+MP&  & &  & & &$\surd$& \\
\hline
 & \multicolumn{7}{c}{iLIDS-VID} \\
\hline
CMC1 & 64.8 & 58.2& 69.1 & 54.9& 72.6$\pm4.8$& 43.6& {\bf77.1$\pm3.7$}\\
CMC5 & 85.5 & 85.7& 88.9 & 77.2& 91.6$\pm2.1$& 68.2&{\bf93.2$\pm2.3$}\\
CMC10& 89.9 & 92.2& 94.0 & 85.7& 96.4$\pm1.1$& 76.5& {\bf97.7$\pm1.2$}\\
CMC20& 95.3 & 98.0& 97.3 & 92.9& 99.1$\pm0.8$& 83.0& {\bf99.4$\pm0.5$}\\
\hline
  & \multicolumn{7}{c}{PRID2011} \\
\hline
CMC1 & 81.6 & 76.9& 86.1 & 75.7& 90.0$\pm2.9$& 39.7& {\bf91.8$\pm2.8$}\\
CMC5 & 96.5 & 95.6& 97.2 & 91.7& 98.8$\pm1.0$& 64.9&{\bf98.4$\pm1.0$}\\
CMC10& 98.3 & 99.1& 99.1 & 96.1& 99.8$\pm0.3$& 76.1& {\bf99.3$\pm0.6$}\\
CMC20& 99.1 & 99.8& 99.6 & 97.9& 100.0$\pm0.0$& 85.7& {\bf99.8$\pm0.4$}\\
\hline
\end{tabular}}

\label{table:ablation}
\end{table}

\subsection{Ablation Study on iLIDS-VID and PRID2011}
\begin{figure}[t]
\centering
\includegraphics[width=1\linewidth]{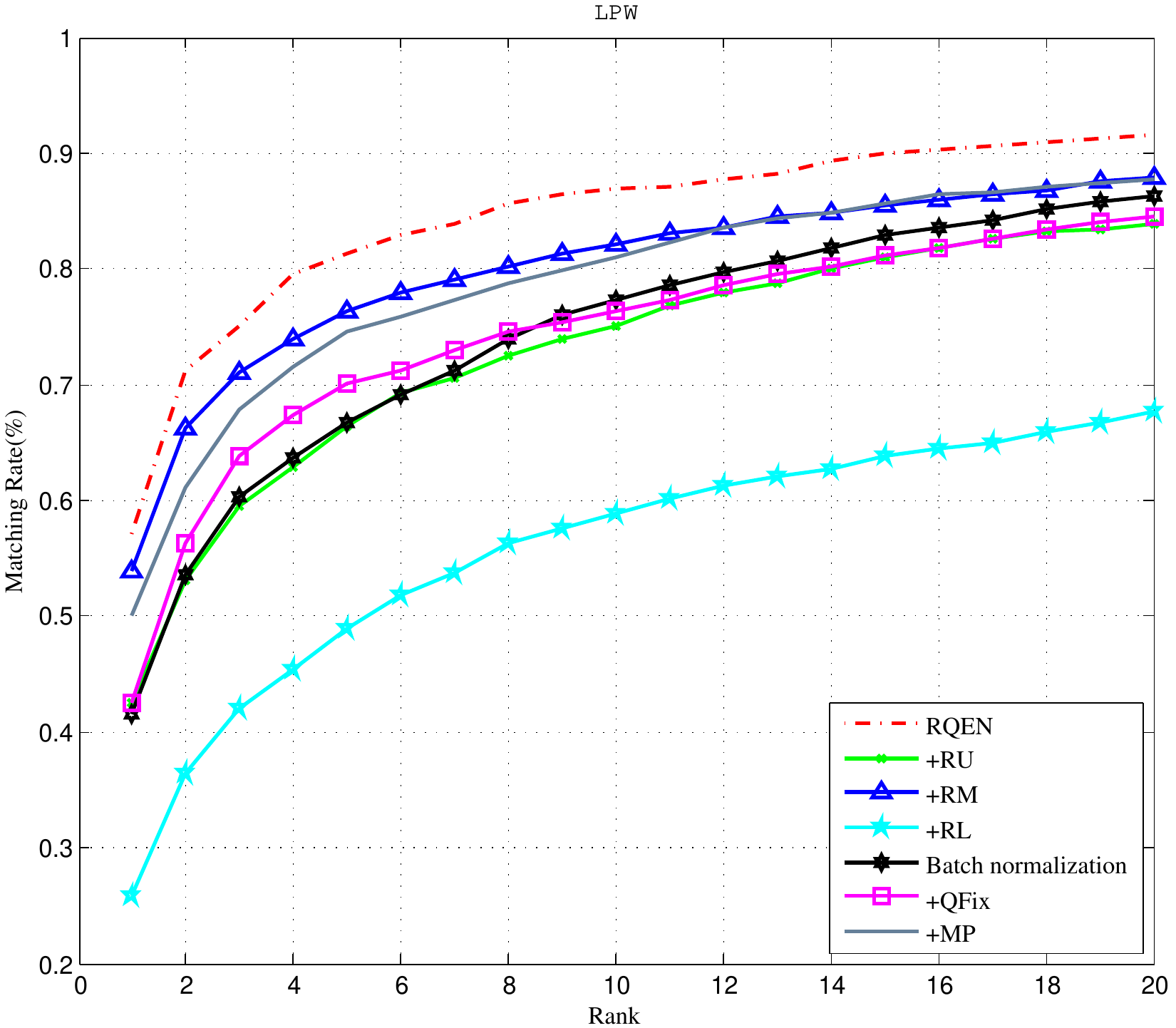}
   \caption{Performance of RQEN, baseline and ablation study experiments on LPW.}
\label{fig:LPW}
\end{figure}

Table~\ref{table:ablation} compares the influence of different units in RQEN. We remark that in this table all results are trained with the same data set and tested using the same configuration.
Method(a) is the GoogLeNet~\cite{szegedy2015going} with batch normalization~\cite{ioffe2015batch} initialized with the ImageNet model and benefited from stronger generalization ability of GoogLeNet with batch normalization, the performance of our baseline has achieved higher accuracy. In method(b),(c) and (d), $+RU$, $+RM$ and $+RL$ represent the features of different regions as mentioned in above and in the result, we can notice that the middle part has a wealth of information which fits our perception that the upper part and middle part of the body plays a better role in identifying a person. In order to verify the effectiveness of region-based quality generation part, we performed two other comparisons method(e) and (f). In the first experiment $+QFix$, we fixed the quality generation unit by setting all the quality scores to 1.
The results have a large variance on PRID2011 and iLIDS-VID because of the limited identities. So the evaluations are repeated $100$ times to get the average accuracy and the standard deviation for (e) and (g). Notice that joint training of image-level and video-level improves performance to a certain extent and with the assist of region-based quality, there is a greater increase in performance.
Considering that RQEN has increased the number of parameters and in order to explore the impact of parameter increase on performance, we configure $+MP$ which has the same number of parameters with RQEN. The branches generated scores is used for generating features and we directly concat the scalar with the feature produced by these branches to be the final feature. Over-fitting of the network caused by insufficient data results in the decrease of performance of method(f) on iLIDS-VID. Method(g) is the configuration of RQEN and all the region proposals and the corresponding region-based quality scores are used. The result of ablation study experiments on large-scale LPW is shown in Figure~\ref{fig:LPW}.

\begin{table}[t]
\centering
\caption{Comparison with related methods on PRID 2011.}
\scalebox{0.8}[1]{
\begin{tabular}{l|c|c|c|c}
\hline
  Matching rate(\%)& \multicolumn{4}{c}{PRID2011} \\
  \hline
  Rank & 1 & 5 & 10 & 20 \\
\hline
RQEN & {\bf91.8$\pm2.8$} & {\bf98.4$\pm1.0$} & {\bf99.3$\pm0.6$} & {\bf99.8$\pm0.4$} \\
\hline
QAN& 90.3 & 98.2 & 99.3 & 100.0 \\
CNN+SRM+TAM&79.4 &94.4 &-&99.3 \\
CNN+XQDA & 77.3 & 93.5 & - & 99.3 \\
CNN+RNN & 70 & 90 & 95 & 97 \\
STA& 64.1 & 87.3 & 89.9 & 92.0 \\
TDL & 56.7 & 80.0 & 87.6 & 93.6 \\
\hline
\end{tabular}}

\label{table:prid2011_performance}
\end{table}

\begin{table}[t]
\centering
\caption{Comparison with related methods on iLIDS-VID.}
\scalebox{0.8}[1]{
\begin{tabular}{l|c|c|c|c}
\hline
  Matching rate(\%)& \multicolumn{4}{c}{iLIDS-VID} \\
  \hline
  Rank & 1 & 5 & 10 & 20 \\
\hline
RQEN & {\bf77.1$\pm3.7$} & {\bf93.2$\pm2.3$} & {\bf97.7$\pm1.2$} & {\bf99.4$\pm0.5$} \\
\hline
QAN& 68.0 & 86.8 & 95.4 & 97.4 \\
CNN+SRM+TAM&55.2&86.5&-&97.0 \\
CNN+XQDA& 53.0 & 81.4 & - & 95.1 \\
CNN+RNN& 58 & 84 & 91 & 96 \\
STA& 44.3 & 71.7 & 83.7 & 91.7 \\
TDL& 56.3 & 87.6 & 95.6 & 98.3 \\
\hline
\end{tabular}}
\label{table:ilids2011}
\end{table}

\begin{table*}[t]
\centering
\caption{Transfer learning ability of LPW. Comparison of performance with different pre-train dataset ImageNet and LPW.}
\begin{tabular}{l|c|c|c|c|c|c|c|c}
\hline
Matching rate(\%)& \multicolumn{4}{c}{PRID 2011}&\multicolumn{4}{c}{iLIDS-VID}  \\
\hline
 Rank & 1 & 5 & 10 & 20 & 1 & 5 & 10 & 20\\
\hline
RQEN+ImageNet & 91.8 & 98.4& \bf{99.3} & 99.8 & 77.1 & 93.2 & 97.7 & \bf{99.4}\\
RQEN+LPW & \bf{93.4} & \bf{98.4} & 99.2 & \bf{100.0} & \bf{80.0} & \bf{94.4} & \bf{98.2} & 99.3\\
\hline
\end{tabular}
\label{table:eva}
\end{table*}

\begin{table}[t]
\centering
\caption{Cross-dataset testing performance on PRID 2011. CD~\cite{hu2014cross} is trained on Shinpuhkan 2014 dataset and other methods are trained on iLIDS-VID. The PRID2011 is used for testing.}
{
\begin{tabular}{l|c|c|c|c}
\hline
Matching rate(\%)& \multicolumn{4}{c}{PRID 2011} \\
\hline
 Rank & 1 & 5 & 10 & 20 \\
\hline
RQEN & {\bf61.8} & {\bf82.6} & {\bf90.4} & {\bf96.1} \\
QAN &34.0 &61.3 & 74.0 &83.1 \\
Batch normalization & 34.3 & 61.2 & 76.4 & 88.8 \\
CNN+RNN& 28 & 57 & 69 & 81 \\
CD~\cite{hu2014cross} & 17 & - & 43 & 52 \\
\hline
\end{tabular}}

\label{table:crossp}
\end{table}

\subsection{Comparison with Published Results}
In the stage of the experiment, results of evaluation are shown in Table~\ref{table:prid2011_performance}, Table~\ref{table:ilids2011}, Table~\ref{table:mars_performance} and Figure~\ref{fig:LPW}. The result is compared with other state-of-the-art methods including CNN+XQDA~\cite{zheng2016mars}, CNN+RNN~\cite{mclaughlin2016recurrent}, STA~\cite{liu2015spatio}, TDL~\cite{you2016top}, CNN+SRM+TAM~\cite{tn2017see}, Reranking~\cite{Zhong_2017_CVPR}, TriNet~\cite{Li_2017_CVPR} and QAN~\cite{yu2017quality} where $Matching rate(\%)$ at Rank i means the accuracy of the matching within the top i gallery classes.

\begin{table}[t]
\centering
\caption{Comparison with state-of-the-art methods on MARS dataset.}
\begin{tabular}{l|c|c|c|c}
\hline
  Matching rate(\%)& \multicolumn{4}{c}{MARS} \\
  \hline
  Rank & 1 & 5& 20 & mAP\\
\hline
our+XQDA+Rerank & 77.83 & 88.84& 94.29 &{\bf71.14}\\
our &73.74 &84.9&91.62&51.70\\
\hline
TriNet+Reranking&{\bf83.03}&{\bf93.69}&{\bf97.63}&66.43\\
IDE+Reranking &73.94&-&-&68.45\\
CNN+SRM+TAM &70.6 &90.0&97.6&50.7\\
CNN+XQDA & 65.0 & 81.1 & 88.9 &49.3 \\
\hline
\end{tabular}
\label{table:mars_performance}
\end{table}

 On PRID 2011 dataset, RQEN increases top-1 accuracy by 1.5\% compared with the state-of-the-art. On iLIDS-VID dataset, RQEN increases top-1 accuracy by 9.1\% compared with the state-of-the-art. Notice that the performance on iLIDS-VID has more improvement because the iLIDS-VID dataset has more images with partial occlusion, deformation, and our model can be effective for selecting the complementary information of the images' regions with high confidence, abandoning the inherent noise of some regions in other frames.
On the large dataset MARS, RQEN achieves the comparable results with the state-of-the-art. Because of the serious misalignment, there is a certain deviation in the region proposals used by RQEN. This situation can affect the performance of the network but the RQEN still has impressive performance benefited from a certain degree of translation invariance of CNN.
On the more challenging LPW, we also configure the ablation study on large-scale dataset LPW. From Figure~\ref{fig:LPW}, we can notice that the RQEN still has stable performance and increase top-1 accuracy by 15.6\% compared with the baseline. $+QM$ has a wealth of information for recognition. In order to verify the effectiveness of part quality generation part, we performed two other comparisons. In the $+QFix$, it has the similar performance with Batch normalization. In the $+MP$, the growth of parameters does bring performance improvements but still not good as RQEN. The experimental results show that the RQEN does improve performance effectively by automatically learning the complementary information between different frames in a sequence.

\subsection{Cross-Dataset Testing}

 Cross-dataset testing is a better way to confirm the validity of our model. Therefore, to better understand how well our proposed system generalizes, we also perform cross-dataset testing.
 For comparing with other methods in cross-dataset evaluation, we adopt the same protocol as~\cite{mclaughlin2016recurrent} and the result is shown in Table~\ref{table:crossp}. We can notice that the RQEN outperforms the average pooling method of baseline and other state-of-the-art.

 \subsection{Transfer Learning Ability of LPW}

We conduct two experiments to study the relationship between LPW and the two small datasets iLIDS-VID and PRID2011. We firstly train RQEN using LPW and then fine tune the RQEN on iLIDS-VID and PRID2011, respectively. Table~\ref{table:eva} shows the result and on both dataset. On iLIDS, fine-tuning from LPW is higher than ImageNet by $~4\%$ in rank-1 accuracy and on PRID2011 1\% higher in rank-1. Note that LPW dataset is of great help to initial training in pedestrian re-identification.

\subsection{Complementary Information Extraction}

\begin{figure}[t]
\centering
\includegraphics[width=1\linewidth]{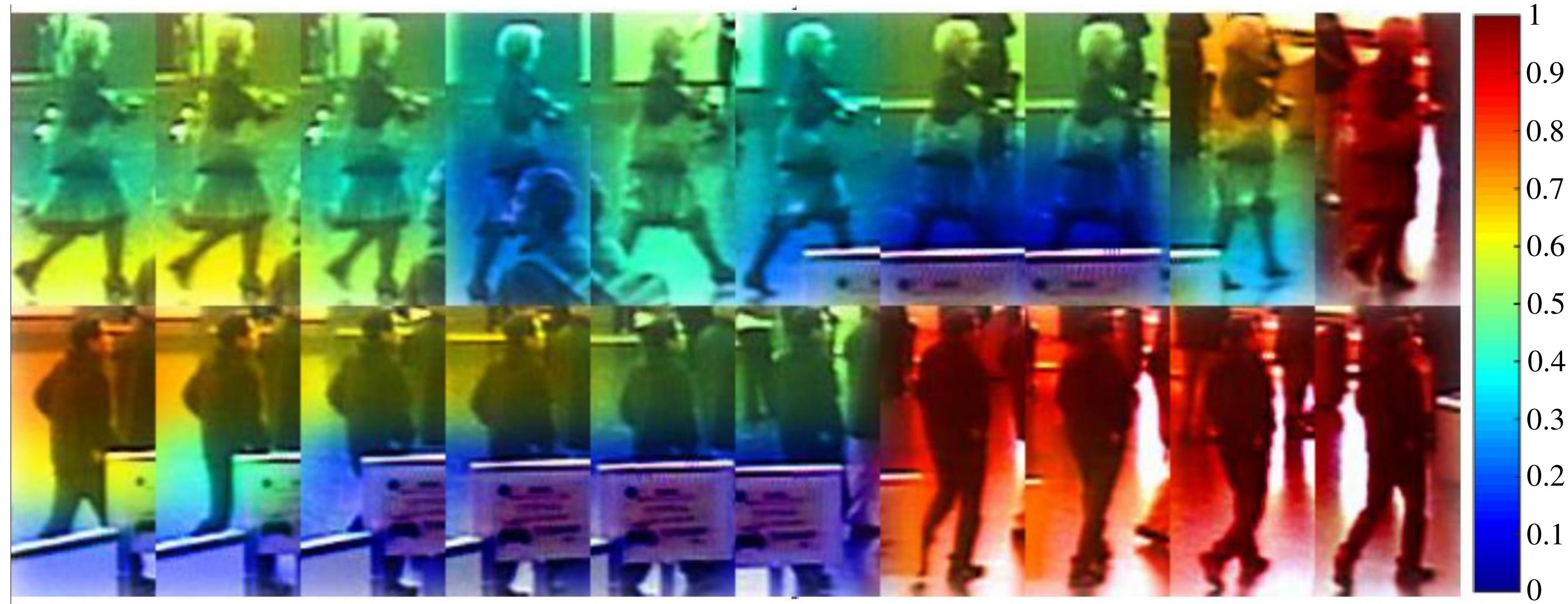}
   \caption{The quality scores of selected images' different parts. The scores in [0,1] are mapped to the colors in the transition from blue to red.}
\label{fig:score}
\end{figure}

We generated the quality scores of the test set using RQEN used for intuitive exploration.
Some partially occluded images are selected for evaluation. Figure~\ref{fig:score} provides a vivid illustration of how the scores reflect the corresponding regional quality.
The qualitiy of images' regions with some noise such as occlusion and blur is poor and the corresponding scores are lower.

Figure~\ref{fig:feat} clearly demonstrates the advantages of RQEN in the aggregation of sequence features. Compared with the average pooling method, it can make more effective use of complementary information in different images' parts of the sequence. The effective information in the region of one frame can make up for missing portions of the same region's information in other frames. The more robust feature representation can be extracted via RQEN with the assist of partial quality scores and get better performance.

\section{Conclusion}

This paper proposes the Region-based Quality Estimation Network (RQEN) for video-based person re-identification. RQEN can learn partial quality of each image and aggregate the images' complementary partial information of different frames in an image sequence. We design an end-to-end training strategy with ingenious gradient design and jointly train the network with classification and verification signal. The RQEN leads to the state-of-the-art results on PRID 2011 and iLIDS-VID. We also propose a large-scale and clean dataset named ``Labeled Pedestrian in the Wild (LPW)'' which contains 7,694 tracklets with over 590,000 images. The dataset, which has occlusion, large posture change and huge age span, is more challenging and suitable for practical use.

\section{Acknowledgements}

This work is supported by the National Natural Science Foundation of China (61472023).

\bibliographystyle{aaai}
\bibliography{egbib}

\begin{thebibliography}{}

\bibitem[\protect\citeauthoryear{Ahmed, Jones, and
  Marks}{2015}]{ahmed2015improved}
Ahmed, E.; Jones, M.; and Marks, T.~K.
\newblock 2015.
\newblock An improved deep learning architecture for person re-identification.
\newblock In {\em CVPR},  3908--3916.

\bibitem[\protect\citeauthoryear{Cheng \bgroup et al\mbox.\egroup
  }{2011}]{cheng2011custom}
Cheng, D.~S.; Cristani, M.; Stoppa, M.; Bazzani, L.; and Murino, V.
\newblock 2011.
\newblock Custom pictorial structures for re-identification.
\newblock In {\em BMVC}, volume~2, ~6.

\bibitem[\protect\citeauthoryear{Cheng \bgroup et al\mbox.\egroup
  }{2016}]{cheng2016person}
Cheng, D.; Gong, Y.; Zhou, S.; Wang, J.; and Zheng, N.
\newblock 2016.
\newblock Person re-identification by multi-channel parts-based cnn with
  improved triplet loss function.
\newblock In {\em CVPR},  1335--1344.

\bibitem[\protect\citeauthoryear{Dikmen \bgroup et al\mbox.\egroup
  }{2010}]{dikmen2010pedestrian}
Dikmen, M.; Akbas, E.; Huang, T.~S.; and Ahuja, N.
\newblock 2010.
\newblock Pedestrian recognition with a learned metric.
\newblock In {\em ACCV},  501--512.
\newblock Springer.

\bibitem[\protect\citeauthoryear{Ding \bgroup et al\mbox.\egroup
  }{2015}]{ding2015deep}
Ding, S.; Lin, L.; Wang, G.; and Chao, H.
\newblock 2015.
\newblock Deep feature learning with relative distance comparison for person
  re-identification.
\newblock {\em Pattern Recognition} 48(10):2993--3003.

\bibitem[\protect\citeauthoryear{Hirzer \bgroup et al\mbox.\egroup
  }{2011}]{hirzer2011person}
Hirzer, M.; Beleznai, C.; Roth, P.~M.; and Bischof, H.
\newblock 2011.
\newblock Person re-identification by descriptive and discriminative
  classification.
\newblock In {\em Scandinavian conference on Image analysis},  91--102.
\newblock Springer.

\bibitem[\protect\citeauthoryear{Hu \bgroup et al\mbox.\egroup
  }{2014}]{hu2014cross}
Hu, Y.; Yi, D.; Liao, S.; Lei, Z.; and Li, S.~Z.
\newblock 2014.
\newblock Cross dataset person re-identification.
\newblock In {\em ACCV},  650--664.
\newblock Springer.

\bibitem[\protect\citeauthoryear{Ioffe and Szegedy}{2015}]{ioffe2015batch}
Ioffe, S., and Szegedy, C.
\newblock 2015.
\newblock Batch normalization: Accelerating deep network training by reducing
  internal covariate shift.
\newblock In {\em Proceedings of the 32nd International Conference on Machine
  Learning}, volume~37,  448--456.

\bibitem[\protect\citeauthoryear{Karanam \bgroup et al\mbox.\egroup
  }{2016}]{resystematic}
Karanam, S.; Gou, M.; Wu, Z.; Rates-Borras, A.; Camps, O.; and Radke, R.~J.
\newblock 2016.
\newblock A systematic evaluation and benchmark for person re-identification:
  Features, metrics, and datasets.
\newblock {\em arXiv preprint arXiv:1605.09653}.

\bibitem[\protect\citeauthoryear{Karanam, Li, and
  Radke}{2015}]{karanam2015sparse}
Karanam, S.; Li, Y.; and Radke, R.~J.
\newblock 2015.
\newblock Sparse re-id: Block sparsity for person re-identification.
\newblock In {\em CVPR},  33--40.

\bibitem[\protect\citeauthoryear{Koestinger \bgroup et al\mbox.\egroup
  }{2012}]{koestinger2012large}
Koestinger, M.; Hirzer, M.; Wohlhart, P.; Roth, P.~M.; and Bischof, H.
\newblock 2012.
\newblock Large scale metric learning from equivalence constraints.
\newblock In {\em CVPR},  2288--2295.
\newblock IEEE.

\bibitem[\protect\citeauthoryear{Li \bgroup et al\mbox.\egroup
  }{2014}]{li2014deepreid}
Li, W.; Zhao, R.; Xiao, T.; and Wang, X.
\newblock 2014.
\newblock Deepreid: Deep filter pairing neural network for person
  re-identification.
\newblock In {\em CVPR},  152--159.

\bibitem[\protect\citeauthoryear{Li \bgroup et al\mbox.\egroup
  }{2017}]{Li_2017_CVPR}
Li, D.; Chen, X.; Zhang, Z.; and Huang, K.
\newblock 2017.
\newblock Learning deep context-aware features over body and latent parts for
  person re-identification.
\newblock In {\em CVPR}.

\bibitem[\protect\citeauthoryear{Liu \bgroup et al\mbox.\egroup
  }{2015}]{liu2015spatio}
Liu, K.; Ma, B.; Zhang, W.; and Huang, R.
\newblock 2015.
\newblock A spatio-temporal appearance representation for viceo-based
  pedestrian re-identification.
\newblock In {\em ICCV},  3810--3818.

\bibitem[\protect\citeauthoryear{Liu, Li, and Wang}{2017}]{liu2017rethinking}
Liu, Y.; Li, H.; and Wang, X.
\newblock 2017.
\newblock Rethinking feature discrimination and polymerization for large-scale
  recognition.
\newblock {\em arXiv preprint arXiv:1710.00870}.

\bibitem[\protect\citeauthoryear{Liu, Yan, and Ouyang}{2017}]{yu2017quality}
Liu, Y.; Yan, J.; and Ouyang, W.
\newblock 2017.
\newblock Quality aware network for set to set recognition.
\newblock In {\em CVPR}.

\bibitem[\protect\citeauthoryear{McLaughlin, Martinez~del Rincon, and
  Miller}{2016}]{mclaughlin2016recurrent}
McLaughlin, N.; Martinez~del Rincon, J.; and Miller, P.
\newblock 2016.
\newblock Recurrent convolutional network for video-based person
  re-identification.
\newblock In {\em CVPR},  1325--1334.

\bibitem[\protect\citeauthoryear{Roth \bgroup et al\mbox.\egroup
  }{2014}]{roth2014mahalanobis}
Roth, P.~M.; Hirzer, M.; K{\"o}stinger, M.; Beleznai, C.; and Bischof, H.
\newblock 2014.
\newblock Mahalanobis distance learning for person re-identification.
\newblock In {\em Person Re-Identification}. Springer.
\newblock  247--267.

\bibitem[\protect\citeauthoryear{Schroff, Kalenichenko, and
  Philbin}{2015}]{schroff2015facenet}
Schroff, F.; Kalenichenko, D.; and Philbin, J.
\newblock 2015.
\newblock Facenet: A unified embedding for face recognition and clustering.
\newblock In {\em CVPR},  815--823.

\bibitem[\protect\citeauthoryear{Szegedy \bgroup et al\mbox.\egroup
  }{2015}]{szegedy2015going}
Szegedy, C.; Liu, W.; Jia, Y.; Sermanet, P.; Reed, S.; Anguelov, D.; Erhan, D.;
  Vanhoucke, V.; and Rabinovich, A.
\newblock 2015.
\newblock Going deeper with convolutions.
\newblock In {\em CVPR},  1--9.

\bibitem[\protect\citeauthoryear{Wang \bgroup et al\mbox.\egroup
  }{2007}]{wang2007shape}
Wang, X.; Doretto, G.; Sebastian, T.; Rittscher, J.; and Tu, P.
\newblock 2007.
\newblock Shape and appearance context modeling.
\newblock In {\em ICCV},  1--8.
\newblock IEEE.

\bibitem[\protect\citeauthoryear{Wang \bgroup et al\mbox.\egroup
  }{2014}]{wang2014person}
Wang, T.; Gong, S.; Zhu, X.; and Wang, S.
\newblock 2014.
\newblock Person re-identification by video ranking.
\newblock In {\em ECCV},  688--703.
\newblock Springer.

\bibitem[\protect\citeauthoryear{Wei \bgroup et al\mbox.\egroup
  }{2016}]{wei2016convolutional}
Wei, S.-E.; Ramakrishna, V.; Kanade, T.; and Sheikh, Y.
\newblock 2016.
\newblock Convolutional pose machines.
\newblock In {\em CVPR},  4724--4732.

\bibitem[\protect\citeauthoryear{Wu \bgroup et al\mbox.\egroup
  }{2016}]{wu2016enhanced}
Wu, S.; Chen, Y.-C.; Li, X.; Wu, A.-C.; You, J.-J.; and Zheng, W.-S.
\newblock 2016.
\newblock An enhanced deep feature representation for person re-identification.
\newblock In {\em WACV},  1--8.
\newblock IEEE.

\bibitem[\protect\citeauthoryear{Xiao \bgroup et al\mbox.\egroup
  }{2016}]{xiao2016learning}
Xiao, T.; Li, H.; Ouyang, W.; and Wang, X.
\newblock 2016.
\newblock Learning deep feature representations with domain guided dropout for
  person re-identification.
\newblock In {\em CVPR},  1249--1258.

\bibitem[\protect\citeauthoryear{Yan \bgroup et al\mbox.\egroup
  }{2016}]{yan2016person}
Yan, Y.; Ni, B.; Song, Z.; Ma, C.; Yan, Y.; and Yang, X.
\newblock 2016.
\newblock Person re-identification via recurrent feature aggregation.
\newblock In {\em ECCV},  701--716.
\newblock Springer.

\bibitem[\protect\citeauthoryear{Yang \bgroup et al\mbox.\egroup
  }{2017}]{yang2016neural}
Yang, J.; Ren, P.; Zhang, D.; Chen, D.; Wen, F.; Li, H.; and Hua, G.
\newblock 2017.
\newblock Neural aggregation network for video face recognition.

\bibitem[\protect\citeauthoryear{Yi \bgroup et al\mbox.\egroup
  }{2014}]{yi2014deep}
Yi, D.; Lei, Z.; Liao, S.; and Li, S.~Z.
\newblock 2014.
\newblock Deep metric learning for person re-identification.
\newblock In {\em ICPR},  34--39.
\newblock IEEE.

\bibitem[\protect\citeauthoryear{You \bgroup et al\mbox.\egroup
  }{2016}]{you2016top}
You, J.; Wu, A.; Li, X.; and Zheng, W.-S.
\newblock 2016.
\newblock Top-push video-based person re-identification.
\newblock In {\em CVPR},  1345--1353.

\bibitem[\protect\citeauthoryear{Zhao, Ouyang, and
  Wang}{2014}]{zhao2014learning}
Zhao, R.; Ouyang, W.; and Wang, X.
\newblock 2014.
\newblock Learning mid-level filters for person re-identification.
\newblock In {\em CVPR},  144--151.

\bibitem[\protect\citeauthoryear{Zheng \bgroup et al\mbox.\egroup
  }{2015}]{zheng2015scalable}
Zheng, L.; Shen, L.; Tian, L.; Wang, S.; Wang, J.; and Tian, Q.
\newblock 2015.
\newblock Scalable person re-identification: A benchmark.
\newblock In {\em ICCV},  1116--1124.

\bibitem[\protect\citeauthoryear{Zheng \bgroup et al\mbox.\egroup
  }{2016}]{zheng2016mars}
Zheng, L.; Bie, Z.; Sun, Y.; Wang, J.; Su, C.; Wang, S.; and Tian, Q.
\newblock 2016.
\newblock Mars: A video benchmark for large-scale person re-identification.
\newblock In {\em ECCV},  868--884.
\newblock Springer.

\bibitem[\protect\citeauthoryear{Zhong \bgroup et al\mbox.\egroup
  }{2017}]{Zhong_2017_CVPR}
Zhong, Z.; Zheng, L.; Cao, D.; and Li, S.
\newblock 2017.
\newblock Re-ranking person re-identification with k-reciprocal encoding.
\newblock In {\em CVPR}.

\bibitem[\protect\citeauthoryear{Zhou \bgroup et al\mbox.\egroup
  }{2017}]{tn2017see}
Zhou, Z.; Huang, Y.; Wang, W.; Wang, L.; and Tan, T.
\newblock 2017.
\newblock See the forest for the trees: Joint spatial and temporal recurrent
  neural networks for video-based person re-identification.
\newblock In {\em CVPR}.

\end{thebibliography}
\end{document}